\documentclass{article}
\usepackage{ijcai18}

\usepackage{times}
\usepackage{xcolor}
\usepackage{soul}
\usepackage[utf8]{inputenc}
\usepackage[small]{caption}
\usepackage{amssymb}
\usepackage{color}
\usepackage{amsmath}
\usepackage{comment}
\usepackage{CJKutf8}
\usepackage{makecell}
\usepackage{subcaption}
\usepackage{bigstrut}
\usepackage{url}
\usepackage{multirow}
\usepackage{graphicx}  %Required
\usepackage{tabularx}  %Required
\title{Translations as Additional Contexts for Sentence Classification}

\author{Reinald Kim Amplayo$^\dag${,}
  Kyungjae Lee$^\dag${,}
  Jinyeong Yeo$^\ddag$ \and
  Seung-won Hwang$^\dag$ \\
  $^\dag$Yonsei University, Seoul, South Korea \\
  $^\ddag$Pohang University of Science and Technology, Pohang, South Korea \\
  {\{rktamplayo, lkj0509, seungwonh\}@yonsei.ac.kr~	
  jinyeo@postech.edu} \\
}

\begin{document}

\maketitle
\begin{abstract}
In sentence classification tasks, additional contexts, such as the neighboring sentences, may improve the accuracy of the classifier.
However, such contexts are {\bf domain-dependent} and thus cannot be used for another classification task with an inappropriate domain.
%where only a single sentence is available for classification.
In contrast, we propose the use of translated sentences
as {\bf domain-free} context 
that is always available regardless of the domain.
We find that naive feature expansion of translations gains only marginal improvements and may decrease the performance of the classifier, due to possible inaccurate translations thus producing noisy sentence vectors.
To this end, we present 
%a neural attention-based 
multiple context fixing attachment (MCFA), a series of modules attached to multiple sentence vectors to fix the noise in the vectors using the other sentence vectors as context. 
%We base the fixing of vectors on two criteria: the sentence's self usability and its relative usability with another sentence. 
We show that our method performs competitively compared to previous
%state of the art 
models, achieving best classification performance on multiple data sets.
%To the best of our knowledge,
We are the first to use translations as domain-free contexts for sentence classification.
\end{abstract}

\setlength{\parskip}{0pt}
\setlength{\abovedisplayskip}{0pt}%
\setlength{\belowdisplayskip}{0pt}%
\setlength{\abovedisplayshortskip}{0pt}%
\setlength{\belowdisplayshortskip}{0pt}%
\setlength{\jot}{0pt}
\setlength{\textfloatsep}{10pt}

\renewcommand{\footnotesize}{\normalsize}

\makeatletter
\renewcommand\paragraph{\@startsection{paragraph}{4}{\z@}%
                                      {\parskip}%{3.25ex \@plus1ex \@minus.2ex}%
                                      {-1em}%
                                      {\normalfont\normalsize\bfseries}}

\section{Introduction}\label{sec:intro}

One of the primary tasks in natural language processing (NLP) is sentence classification, where given a sentence (e.g. a sentence of a review) as input, we are tasked to classify it into one of multiple classes (e.g. into positive or negative). This task is important as it is widely used in almost all subareas of NLP such as sentiment classification for sentiment analysis \cite{Pang2007OpinionMA} and question type classification for question answering \cite{li2002learning}, to name a few. While past methods require feature engineering, recent methods enjoy neural-based methods to automatically encode the sentences into low-dimensional dense vectors \cite{Kim2014ConvolutionalNN,Joulin2017BagOT}. 
%These vectors are then used as input features to train a classifier. 
Despite the success of these methods, the major challenge in this task is that extracting features from a single sentence limits the performance.

%modeling and classification. The task is widely used in almost all subareas of NLP such as sentiment analysis \cite{Pang2007OpinionMA} and paraphrase identification \cite{socher2011dynamic}. The task starts by representing the sentences into numerical vectors and uses these vectors as input features to train a classifier to classify the sentences into one of the multiple classes (e.g. into sentiment classes). 
%One basic representation technique is the Bag-of-Words model \cite{Wang2012BaselinesAB}. Other 
%Recent methods widely include neural-based methods that insert non-linearity for better representation of the sentence \cite{Kim2014ConvolutionalNN,Joulin2017BagOT}.

%A limitation of sentence classification is that a data instance only contains one sentence as context to create a feature vector.
To overcome this limitation, recent works attempted to augment different kinds of features to the sentence, such as the neighboring sentences \cite{lin2015hierarchical} 
%, the document containing the sentence \cite{huang2012improving}, 
and the topics of the sentences \cite{zhao2017topic}.
However, these methods used {\bf domain-dependent} contexts that are only effective when the domain of the task is appropriate.
For one thing,
%have several disadvantages. 
neighboring sentences 
%and the full document 
may not be available in some tasks such as question type classification. %\cite{grbovic2015context}. 
Moreover, topics 
%can be 
inferred using topic models 
%however they 
may produce less useful topics when the data set is domain-specific such as movie review sentiment classification \cite{mimno2011optimizing}.
%These 
%domain-sensitive 
%contexts are only effective when the domain of the task is appropriate.
%In contrast, \textbf{domain-free additional contexts} are more superior.

%\begin{figure}[t]
%\centering
\begin{figure}[t]
	\centering
	\includegraphics[width=0.47\textwidth]{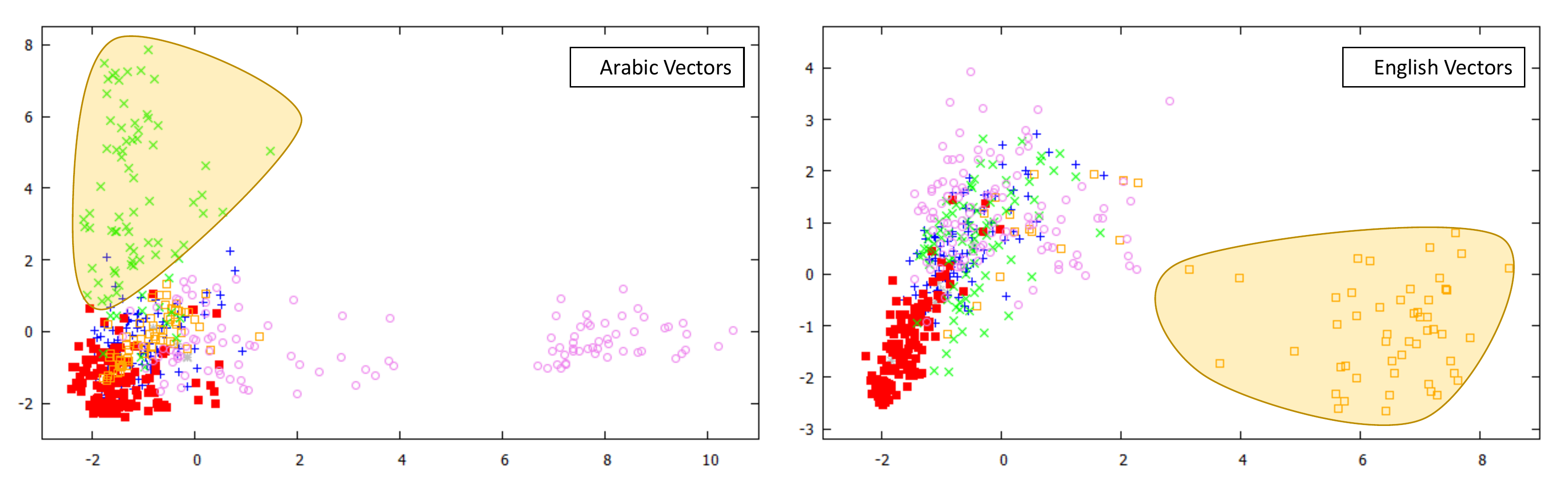}
	\caption{PCA visualizations of unaltered sentence vectors on TREC data set, where each language is effective for a specific class,  highlighted using a yellow circle.}
	\label{fig:trecvecs}
\end{figure}
%\caption{PCA visualization of unaltered sentence vectors encoded using CNN. The colors represent the class of the sentence.}
%\end{figure}

In this paper, 
%we propose to use \textbf{domain-free additional contexts}, contexts that are always usable, self to the domain.
%Specifically, 
we propose the usage of translations as compelling and effective \textbf{domain-free} contexts, or contexts that are %always useful and 
always available no matter what the task domain is. We observe two opportunities when using translations.
%We have three hypotheses:
%\begin{itemize}
%\item A classification task (or its subtask) is more effective in one language than another.
%\item The given sentence can be less ambiguous in one language than another.
%\item Translation generates both signals and noises, and noises should be controlled.
%\end{itemize}
\begin{CJK}{UTF8}{}
	\CJKfamily{mj}
%\paragraph{(1) Diverse Effectiveness}

First, each language has its own linguistic and cultural characteristics that may contain different signals to effectively classify a specific class.
%one language is more effective than another in 
%classifying one class from another.
Figure \ref{fig:trecvecs} contrasts the sentence vectors of
the original English sentences and their Arabic-translated sentences in the question type classification task.
A yellow circle signifies a clear separation of a class.
%is clearly separated in one language but not in another.
For example, 
the green class, or the numeric question type, is circled in the Arabic space as it is clearly separated from other classes, while such separation
%distinction 
cannot be observed
in English. Meanwhile, location type questions (in orange)
%classes
are better classified in English.

%\paragraph{(2) Ambiguity Resolution}
Second, the original 
%given
sentences may include language-specific ambiguity, which may be resolved when presented with its translations.
%translations automatically resolve ambiguities in the original sentence. 
Consider the example English sentence ``\textit{The movie is terribly amazing}'' for the sentiment classification task. In this case, 
\textit{terribly}
%is an ambiguous word that
%because it 
can be used in both 
%the 
positive and negative sense, 
%and 
thus introduces ambiguity in the sentence. When translated to Korean,
% using Google Translate,
it becomes ``\textit{영화는 대단히 훌륭합니다}'' which means ``\textit{The movie is greatly magnificent}'', removing the ambiguity.
%when the original sentence contains ambiguous expressions, the translated sentence may be able to remove the ambiguous expressions and make such expressions easier to classify.
\end{CJK}

%Lastly, t
The above two observations hold only when translations are supported for (nearly) arbitrary language pairs with sufficiently high quality.
Thankfully, translation services (e.g. Google Translate)
%\footnote{\url{https://translate.google.com/}}) provide readily available translation systems for many pairs of source and target languages.
%, thus translations are always available for all kinds of domains and languages.
Moreover, recent research on neural machine translation (NMT) \cite{bahdanau2014neural} improved the efficiency
%of the translation models 
and even enabled 
%the ability to do 
zero-shot translation \cite{johnson2016google} 
of models
for languages with no parallel data.
This provides an opportunity to leverage on as many languages as possible to any domain, providing a much wider context compared to the limited contexts provided by past studies.

\begin{figure}[t]
	\centering
	\includegraphics[width=0.47\textwidth]{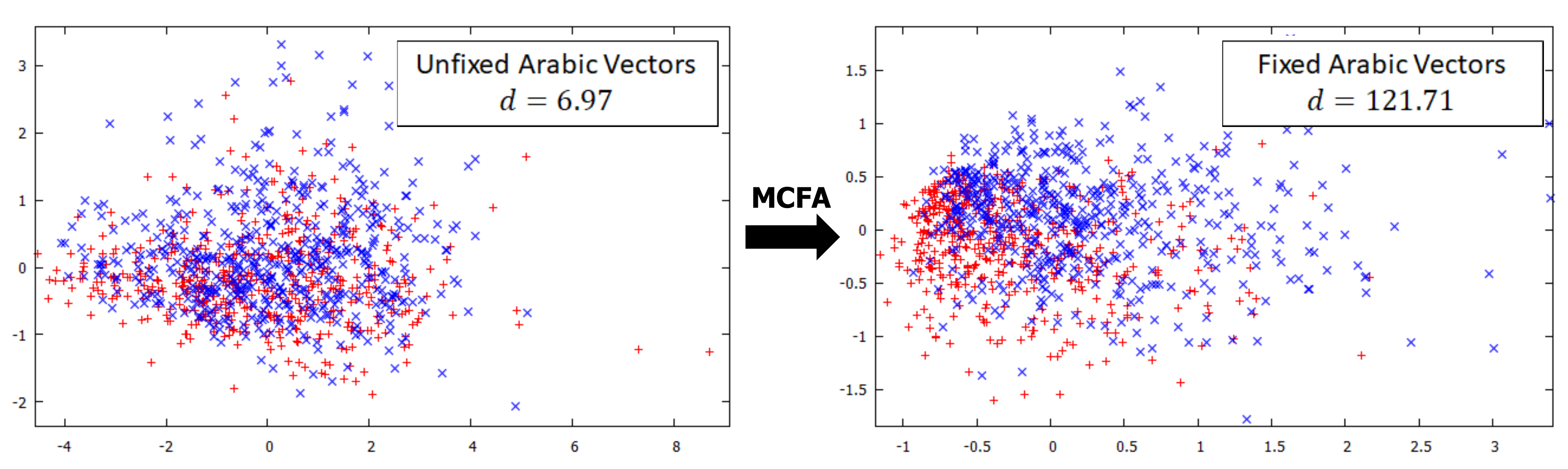}
	\caption{PCA visualizations of unaltered sentence vectors (left) and the corresponding MCFA-altered vectors (right)  on the MR data set.
	$d$ is the Mahalanobis distance between the two class clusters.
	%Fixing is done by relocating the vectors for better separation.
	%where classes in a translated space (i.e. Arabic) are indistinguishable.
	}
	\label{fig:mrvecs}
\end{figure}

However, despite the maturity of translation,
%One way to use these contexts is to 
naively concatenating their vectors to the original sentence vector
%However, translations of the machine translation system are still far from perfect and 
may introduce more noise 
%to the classifier
than signals.
%that may cause adverse effects to its performance. This phenomenon is better seen visually in 
The unaltered translation space on the left of Figure \ref{fig:mrvecs} shows an example
%unaltered translated space, 
where translation noises make the two classes indistinguishable.
%where the vector space contains a lot of noise that makes it hard for the classifier to classify the sentiment of the sentence. 
%Applying regularization techniques may help alleviate the problem by weakening the importance of some dimensions of the vectors, however this method does not fully use the potential of the additional contexts and could also render them useless if all the weights are weakened.

%are still not at par with human translation accuracy. This results to inaccurate translations for some sentences which are harder to translate. This further results to an incorrect feature vector representation and to the decrease in the classifier accuracy.

In this paper, we propose a method to mitigate the possible problems when using translated sentences as context based on the following observations. Suppose there are two translated sentences $a$ and $b$ with slight errors. We posit that $a$ can be used to fix $b$ when $a$ is used as a context of $b$, and vice versa\footnote{Hereon, we mean to ``fix'' as to ``correct, repair, or alter.''}. Revisiting the example above, to fix the vector of the English sentence ``\textit{The movie is terribly amazing}'', we use the Korean translation to move the vector towards the location where the vector ``\textit{The movie is greatly magnificent}'' is.

Based on these observations, we present a neural attention-based multiple context fixing attachment (MCFA). MCFA is a series of modules that uses all the sentence vectors (e.g. Arabic, English, Korean, etc.) as context to fix a sentence vector (e.g. Korean). 
%Through MCFA, we put weights to the sentences and aggregate them into one context vector. Here, the weights correspond to the language's usability and its relative importance with the other languages. The aggregated context vector is then used to create a gate vector that is used for fxing.
Fixing the vectors is done by selectively moving the vectors to a location in the same vector space that better separates the class, as shown in Figure~\ref{fig:mrvecs}.
%to the appropriate location in the same vector space. We do this correction for all languages creating an environment where all languages help each other for correction.
Noises from translation may cause adverse effects to the vector
\textbf{itself} (e.g. when a noisy vector is directly used for the task) and \textbf{relatively} to other vectors (e.g. when a noisy vector is used to fix another noisy vector).
%during the vector correction phase.
MCFA computes two sentence
usability metrics to control the noise when fixing vectors:
(a) \textbf{self usability} $\rho_i(a)$ weighs the confidence of using sentence $a$ in solving the task.
(b) \textbf{relative usability} $\rho_r(a, b)$ weighs the confidence of using sentence $a$ in fixing sentence $b$.

%usability modules to control the noise and solve these problems.
Listed below are the three main strengths of the MCFA attachment.
(1) 
%\textbf{Adaptable to any sentence encoders}: 
MCFA is attached
    %an attachment 
    after encoding the sentence, 
    %representation model, 
    which makes it widely adaptable to other models.
(2) 
%\textbf{Expandable to arbitrarily many languages}: 
MCFA is extensible and improves the accuracy as the number of translated sentences increases.
    %, unlike naive concatenation where the accuracy decreases.
(3) 
%\textbf{Interpretable vector corrections}: 
MCFA 
    %does not project the sentence vectors into another vector space and instead just 
    moves the vectors inside the same space, thus preserves the meaning of vector dimensions.
Results show that a convolutional neural network (CNN) attached with MCFA significantly improves the classification performance of CNN, achieving state of the art performance over multiple data sets. 
%We also show qualitative analysis of our model and provide visualization as to how MCFA fixes vectors based on the 
%sentence usability metrics during the vector correction phase\footnote{The code we use in this paper is publicly shared: \url{http://anonymous.link}.}.
%self usabilities of context vectors and their sentence similarities with the current vector to be altered.

\section{Preliminaries}

%In this section, we define the problem on how to use translations as additional context. We also discuss the base sentence representation model we use in this paper and two naive extensions to utilize translated sentences as context. Note that since our proposed attachment is orthogonal to the model, other better performing supervised sentence encoding techniques \cite{ma2015dependency,zhang2016dependency} can also be used. We use a common model for simplicity.

\begin{figure*}[t]
    \centering
    \begin{subfigure}[t]{0.45\textwidth}
        \centering
        \includegraphics[width=\textwidth]{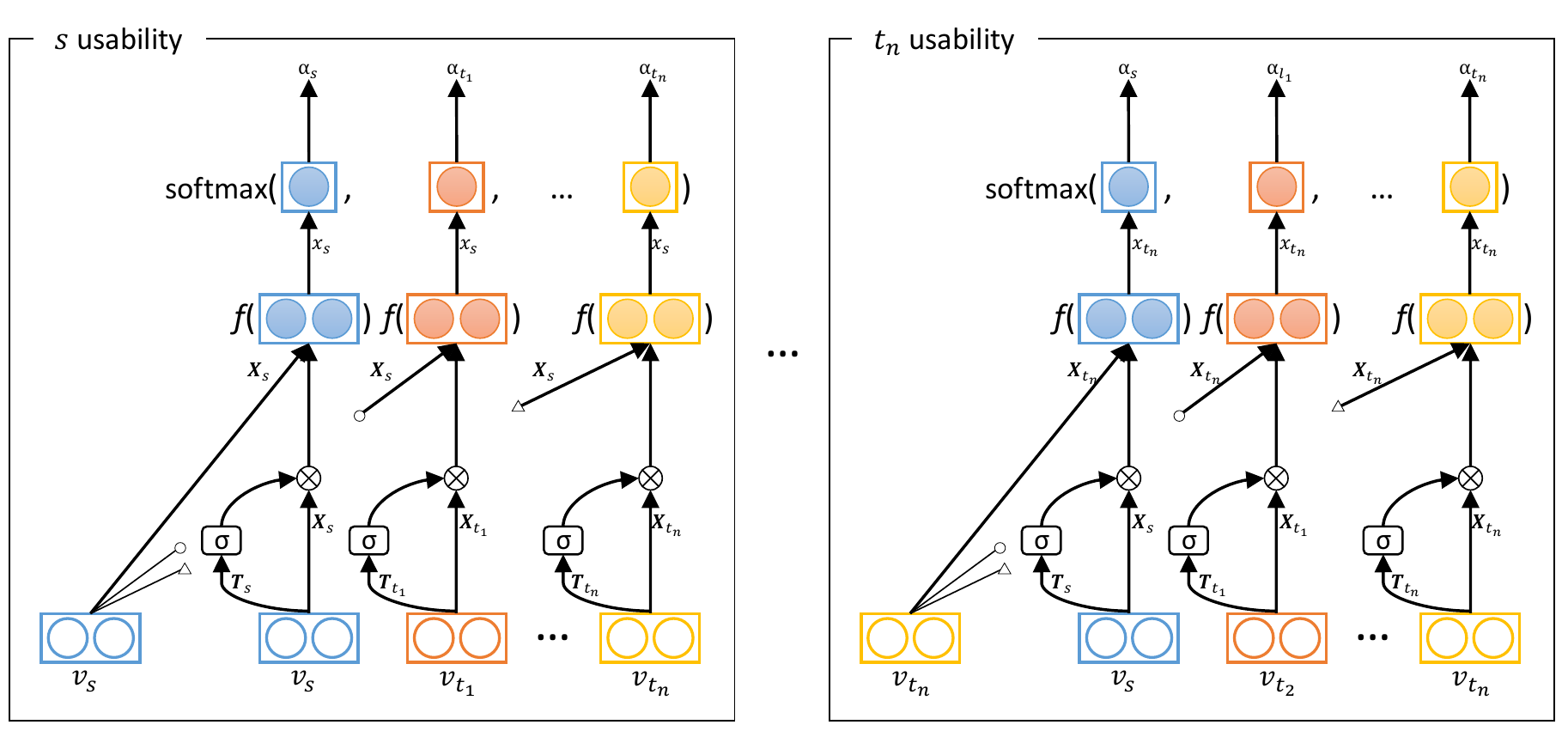}
        \caption{Self and relative usability modules}
        \label{fig:sga}
    \end{subfigure}
    ~ 
    \begin{subfigure}[t]{0.45\textwidth}
        \centering
        \includegraphics[width=\textwidth]{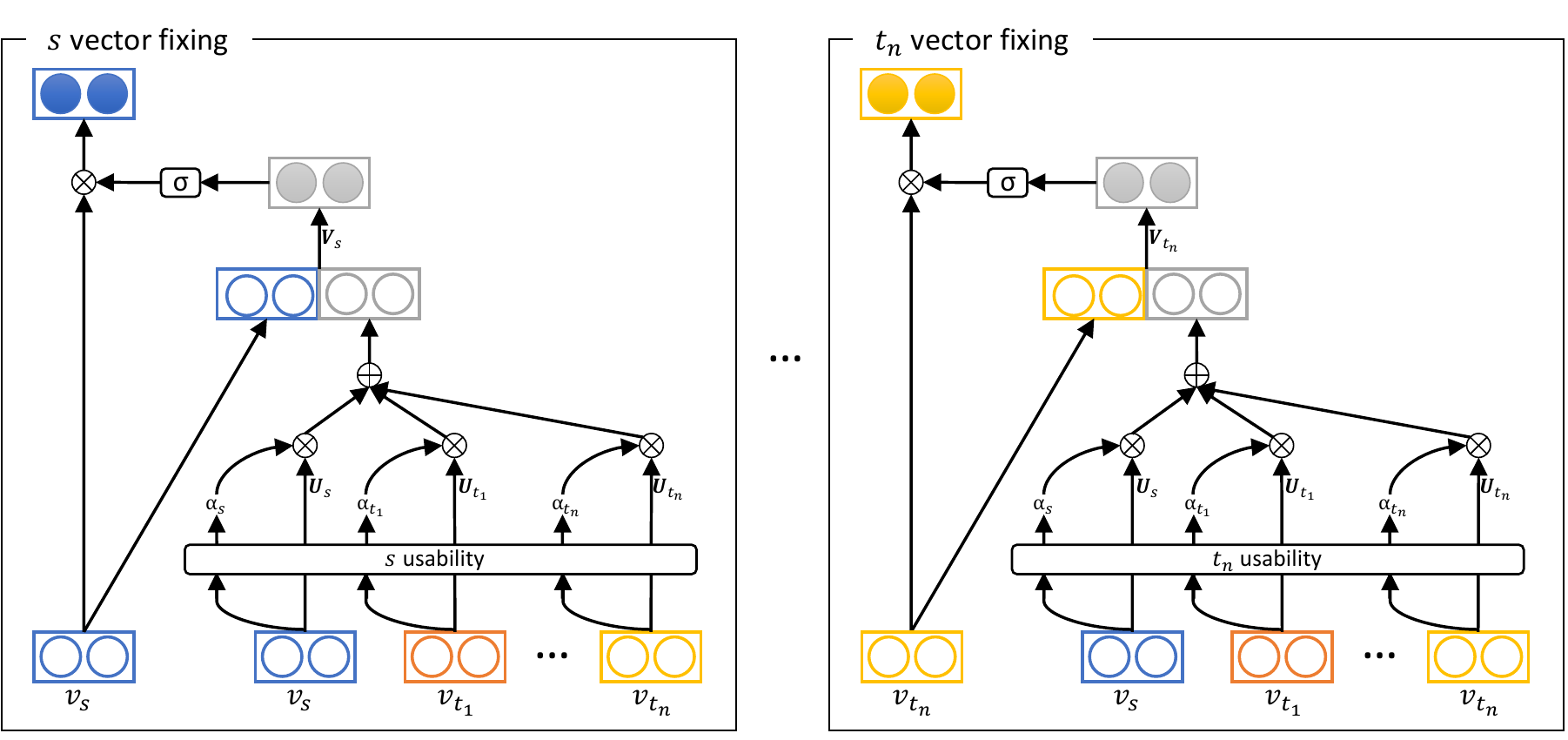}
        \caption{Vector fixing module}
        \label{fig:vf}
    \end{subfigure}
    \caption{Full architecture of the MCFA attachment. An arrow marked with a variable is a matrix multiplication of the vector and the variable. An arrow without a variable simply carries the previous element to the next element.}
    \label{fig:arch}
\end{figure*}

\subsection{Problem: Translated Sentences as Context}

In this paper, the ultimate task that we solve is the sentence classification task where given a sentence and a list of classes, one is task to classify which class (e.g. positive or negative sentiment) among the list of classes does the sentence belong. However, the main challenge that we tackle is the task on how to utilize translated sentences as additional context in order to improve the performance of the classifier.
Specifically, the problem states: given the original sentence $s$, the goal is to use $t_1, t_2, ..., t_n$, or sentences in other languages which are translated from $s$, as additional context.

\paragraph{Base Model: Convolutional Neural Network.}

The base model used is the convolutional neural network (CNN) for sentences \cite{Kim2014ConvolutionalNN}. It is a simple variation of the original CNN for texts \cite{collobert2011natural} to be used on sentences. Let $\mathbf{x}_i \in \mathbb{R}^d$ be the $d$-dimensional word vector of the $i$-th word in a sentence of length $n$. A convolution operation involves applying a filter matrix $\mathbf{W} \in \mathbb{R}^{h \times d}$ to a window of $h$ words and producing a new feature vector $c_i$ using the equation $c_i = f([\mathbf{x}_i; ...; \mathbf{x}_{i+h-1}]^\top \mathbf{W}+b)$, where $b$ is a bias vector and $f(.)$ is a non-linear function.
By doing this on all possible windows of words we produce a feature map $\mathbf{c} = [c_1, c_2, ...]$. We then apply a max-over-time pooling operation \cite{collobert2011natural} over the feature map and take the maximum value as the feature vector of the filter. We do this on all feature vectors and concatenate all the feature vectors to obtain the final feature vector $\mathbf{v}$. We can then use this vector as input features to train a classifier such as logistic regression.
We use CNN to create sentence vectors for all sentences $s, t_1, t_2, ..., t_n$. From here on, we refer to these vectors as $\mathbf{v}_s, \mathbf{v}_{t_1}, \mathbf{v}_{t_2}, ..., \mathbf{v}_{t_n}$, respectively. We refer to them collectively as $\mathbb{V}$.

\paragraph{Baseline 1: Naive Concatenation.}

A simple method in order to use the translated sentences as additional context is to naively concatenate their vectors with the vector of the original sentence. That is, we create a wide vector $\mathbf{\hat{v}} = [\mathbf{v}_s; \mathbf{v}_{t_1}; ...; \mathbf{v}_{t_n}]$, and use this as the input feature vector of the sentence to the 
%logistic regression
classifier.
This method works fine if the translated sentences are translated properly. However, sentences translated using machine translation models usually contain incorrect translation. In effect, this method will have adverse effects on the overall performance of the classifier. This will especially be very evident if the number of additional sentences increases.

\paragraph{Baseline 2: L2 Regularization.}

In order to alleviate the problems above, we can use L2 regularization to automatically select useful features by weakening the appropriate weights. %This is done by inserting the L2 loss of the weight matrix, $\lambda ||w||^2$, to the loss function. 
The main problem of this method occurs when almost all of the weights coming from the vectors of the translated sentence are weakened. This leads to making the additional context vectors useless and to having a similar performance when there are no additional context. Ultimately, this method does not make use of the full potential of the additional context.

\section{Model}

To solve the problems of the baselines discussed above, we introduce an attention-based neural multiple context fixing attachment (MCFA)\footnote{The code we use in this paper is publicly shared: \url{https://github.com/rktamplayo/MCFA}}, a series of modules attached to the sentence vectors $\mathbb{V}$. MCFA attachment is used to fix the sentence vectors, by slightly modifying the per-dimension values of the vector, before concatenating them into the final feature vector. The sentence vectors are altered using other sentence vectors as context (e.g. $\mathbf{v}_{t_1}$ is altered using $\mathbf{v}_s, \mathbf{v}_{t_2}, ..., \mathbf{v}_{t_n}$). This results to moving the vectors %from one place to another
in the same vector space.
%without introducing non-linearity. 
The full architecture is 
%graphically 
shown in Figure \ref{fig:arch}.
%In the next sections, we discuss the subparts of MCFA attachment: the self usability module, the relative usability module, and the vector fixing module. 
%We use these modules to fix each sentence vector.

\subsection{Self Usability Module}

To fix a source sentence vector\footnote{Hereon, we say that $\mathbf{v}_k$ is a \textit{source sentence vector} if $\mathbf{v}_k$ is the current vector to be altered.}, we use the other sentence vectors as guide to know which dimensions to fix and to what extent do we need to fix them. However, other vectors might also contain errors which may reflect to the fixing of the source sentence vector. In order to cope with this, we introduce self usability modules. A self usability module contains the \textbf{self usability} of the vector
%to be used as context. \textbf{Self usability} 
$\rho_i(a)$, which measures how confident sentence $a$ is for the task at hand. For example, an ambiguous sentence (e.g. ``\textit{The movie is terribly amazing}'') may receive a low self usability, while a clear
and definite 
sentence (e.g. ``\textit{The movie is very good}'') may receive a high self usability.
%much we should trust this vector based on how well it was translated, in terms of probability.

Mathematically, we calculate the self usability of the vector $\mathbf{v}_i$ of sentence $i$, denoted as $\rho_{i}(\mathbf{v}_i)$, using the equation $\rho_i(\mathbf{v}_i) = \sigma(\mathbf{v}_i^\top \mathbf{T}_i)$, where $\mathbf{T}_i \in \mathbb{R}^{d \times 1}$ is a matrix to be learned. The produced value is a single real number from 0 to 1.
We pre-calculate the self usability of all sentence vectors $\mathbf{v}_i \in \mathbb{V}$. These are used in the next module, the relative usability module.

\subsection{Relative Usability Module}

\textbf{Relative usability} $\rho_r(a, b)$ measures how useful $a$ can be when fixing $b$, relative to other sentences. There are two main differences between $\rho_i(a)$ and $\rho_r(a, b)$. First, $\rho_i(a)$ is calculated before $a$ knows about $b$ while $\rho_r(a, b)$ is calculated when $a$ knows about $b$. Second, $\rho_r(a, b)$ can be low even though $\rho_i(a)$ is not. This means that $a$ is not able to help in fixing the wrong information in $b$.
%The attention mechanism \cite{bahdanau2014neural} is first introduced in neural machine translation in order to automatically align words from different languages. Recently, different kinds of attention mechanisms are widely used for classification \cite{yang2016hierarchical} to give appropriate weights to different context vectors.
Here, we extend the additive attention module \cite{bahdanau2014neural} and use it as a method to calculate the relative usability of two sentences of different languages. To better visualize the original attention mechanism, we present the equations below.
\begin{align}
    \label{eq:addattscore}
    e_i &= u^\top tanh( s^\top \mathbf{W}+t_i^\top \mathbf{U} ) \\
    \alpha_i &= \frac{exp(e_i)}{\sum_{j \in T} exp(e_j)}
\end{align}
One major challenge in using the attention mechanism in our problem is that the sentence vectors do not belong to the same vector space. Moreover, one characteristic of our problem is that the sentence vectors can be both a source and a context vector (e.g. $\mathbf{v}_s$ can be both $s$ and $t_i$ in Equation \ref{eq:addattscore}). Because of these, we cannot directly use the additive attention module. We extend the module such that (1) each sentence vector $\mathbf{v}_k$ has its own projection matrix $\mathbf{X}_k \in \mathbb{R}^{d \times d}$, and (2) each projection matrix $\mathbf{X}_k$ can be used as projection matrix of both the source (e.g. when sentence $k$ is the current source) and the context vectors. Finally, we incorporate the self usability function $\rho_i(\mathbf{v}_k)$ to reflect the self usability of a sentence. Ultimately, the relative usability denoted as $\rho_r(\mathbf{v}_i, \mathbf{v}_j)$ is calculated using the equations below, where $\times$ is the multiplication of a vector and a scalar through broadcasting.
\begin{align}
    e(\mathbf{v}_i, \mathbf{v}_j) &= x^\top tanh( \mathbf{v}_i^\top \mathbf{X}_i+\mathbf{v}_j^\top \mathbf{X}_j \times \rho_i(\mathbf{v}_j) ) \\
    \rho_r(\mathbf{v}_i, \mathbf{v}_j) &= \frac{exp(e(\mathbf{v}_i, \mathbf{v}_j))}{\sum_{\mathbf{v}_k \in \mathbb{V}} exp(e(\mathbf{v}_i, \mathbf{v}_k))}
\end{align}
\subsection{Vector Fixing Module}

The vector fixing module applies the attention weights to the sentence vectors and creates an integrated context vector. We then use this vector alongside with the source sentence vector to create a weighted gate vector. The weighted gate vector is used to determine to what extent should a dimension of the source sentence vector be altered.

The common way to apply the attention weights to the context vectors and create an integrated context vector $c_i$ is to directly do weighted sum of all the context vectors.
%, as shown in the equation below.
%
%\begin{equation}
%    c_i = \sum_{\mathbf{v_k} \in \mathbb{V}} \alpha_{ik} \mathbf{v}_k
%\end{equation}
%
However, this is not possible because the context vectors are not on the same space. Thus, we use a projection matrix $\mathbf{U}_k \in \mathbb{R}^{d \times d}$ to linearly project the sentence vector $\mathbf{v}_k$ to transform the sentence vectors into a common vector space. The integrated context vector $c_i$ is then calculated as
$c_i = \sum_{\mathbf{v}_k \in \mathbb{V}} \rho_r(\mathbf{v}_i, \mathbf{v}_k) \mathbf{v}_k^\top \mathbf{U}_k$.

Finally, we construct a weighted gate vector $w_k$ and use it to fix the source sentence vectors using the equations below, where $\mathbf{V}_k \in \mathbb{R}^{2d \times d}$ is a trainable parameter and $\otimes$ is the element-wise multiplication procedure. The weighted gate vector is a vector of real numbers between 0 and 1 to modify the intensity of per-dimension values of the sentence vector.
This causes the vector to move in the same vector space towards the correct direction.
\begin{equation}
    w_k = \sigma([\mathbf{v_k}; c_k]^\top \mathbf{V}_k)
\end{equation}
\begin{equation}
    \hat{\mathbf{v}}_k = \mathbf{v}_k \otimes w_k
\end{equation}
An alternative approach to do vector correction is using a residual-style correction, where instead of multiplying a gate vector, a residual vector \cite{he2016deep} is added to the original vector. However, this approach makes the correction not interpretable; it is hard to explain what does adding a value to a specific dimension mean.
One major advantage of MCFA is that the corrections in the vectors are interpretable; the weights in the gate vector correspond to the importance of the per-dimension features of the vector.
The altered vectors $\hat{\mathbf{v_s}}, ..., \hat{\mathbf{v_{t_n}}}$ are then concatenated and fed directly as an input vector to the logistic regression classifier for training.

\section{Experiments}

%In this section, we study the empirical performance of the MCFA attachment on four benchmark data sets for sentence classification and compare it to other competing models.
%We also show interpretations of the modules presented above by providing visualizations and examples.

\subsection{Experimental Setting}

We test our model on four different data sets as listed below and summarized
%. We also show the summary of statistics of the data sets 
in Table \ref{tab:data}.
(a) \textbf{MR}\footnote{\url{https://www.cs.cornell.edu/people/pabo/movie-review-data/}} \cite{pang2005seeing}: 
%Movie reviews where each data instance is a sentence. 
Movie reviews data where the task is to classify whether the review sentence has positive or negative sentiment.
(b) \textbf{SUBJ} \cite{pang2004sentimental}: Subjectivity data where the task is to classify whether the sentence is subjective or objective.
(c) \textbf{CR}\footnote{\url{http://www.cs.uic.edu/~liub/FBS/sentiment-analysis.html}} \cite{hu2004mining}: Customer reviews where
%each data instance is a review of a certain product.
    %(total of 14 products). 
The task is to classify whether the review sentence is positive or negative.
(d) \textbf{TREC}\footnote{\url{http://cogcomp.cs.illinois.edu/Data/QA/QC/}} \cite{li2002learning}: TREC question data set 
%where given a question, 
the task is to classify the type of question.
%is it (total of six question types).

\begin{table}[!t]
	\scriptsize
  \centering
    \begin{tabular}{|c|c|c|c|c|}
    \hline
    \textbf{Data set} & $c$ & $|w|$ & $M$  & \textit{Test} \\
    \hline
    MR    & 2     & 20    & 10662 & CV \\
    SUBJ  & 2     & 19    & 10000 & CV \\
    CR    & 2     & 23    & 3775  & CV \\
    TREC  & 6     & 10    & 5952  & 500 \\
    \hline
    \end{tabular}%
  \caption{Statistics of the four data sets used in this paper. $c$: number of target classes. $|w|$: average number of words. $M$: number of data instances. \textit{Test}: size of the test data, if available. If not, we use 10-fold cross validation (marked as CV) with random split.}
  \label{tab:data}%
\end{table}%

All our data sets are in English. For the additional contexts, we use ten other languages, selected based on their diversity and their performance on prior experiments: Arabic, Finnish, French, Italian, Korean, Mongolian, Norwegian, Polish, Russian, and Ukranian. We translate the data sets using Google Translate. Tokenization is done using the polyglot library\footnote{\url{https://pypi.python.org/pypi/polyglot}}. We experiment on using only one additional context ($N=1$) and using all ten languages at once ($N=10$). For $N=1$, we only show the accuracy of the best classifier for conciseness.

For our CNN, we use rectified linear units and three filters with different window sizes $h=3,4,5$ with $100$ feature maps each, following \cite{Kim2014ConvolutionalNN}. For the final sentence vector, we concatenate the feature maps to get a 300-dimension vector. We use dropout \cite{srivastava2014dropout} on all non-linear connections with a dropout rate of 0.5. We also use an $l_2$ constraint of 3, following
%. These values are based on the CNN paper 
\cite{Kim2014ConvolutionalNN} for accurate comparisons. We use FastText pre-trained vectors\footnote{\url{https://github.com/facebookresearch/fastText/blob/master/pretrained-vectors.md}} \cite{bojanowski2016enriching} for all our data sets and their corresponding additional context.
%FastText provides pre-trained vectors for 294 languages trained on Wikipedia. These vectors have 300 dimensions.
During training, we use mini-batch size of 50. Training is done via stochastic gradient descent over shuffled mini-batches with the Adadelta update rule. We perform early stopping using a random $10\%$ of the training set as the development set.

We present several competing models, listed below to compare the performance of our model. (A) Aside from the base model (\textbf{CNN}) \cite{Kim2014ConvolutionalNN}, we use Dependency-based CNN (\textbf{Dep-CNN}) \cite{ma2015dependency}, which parses the sentences first and does convolution on ancestor paths and Dependency-sensitivity CNN (\textbf{DSCNN}) \cite{zhang2016dependency}, which uses LSTM to capture dependency information within each sentence; (B) \textbf{AdaSent} \cite{zhao2015self} adopts a hierarchical structure, where consecutive levels are connected through gated recursive composition of adjacent segments, and feeds the hierarchy as a multi-scale representation through a gating network;
(C) Topic-aware Convolutional Neural Network (\textbf{TopCNN}) \cite{zhao2017topic} uses topics as additional contexts and changes the CNN architecture. TopCNN uses two types of topics: word-specific topic and sentence-specific topic; and
(D) \textbf{CNN+B1} and \textbf{CNN+B2} are the two baselines presented in this paper. %, which naively concatenate the sentence vectors of the original sentence and the translated sentences, and additionally uses L2 regularization for feature selection. 

We do not show results from RNN models because they were shown to be less effective in sentence classification in our prior experiments.
For models with additional context, we further use an ensemble classification model using a commonly used method by averaging the class probability scores generated by the multiple variants (in our model's case, $N=1$ and $N=10$ models), following \cite{zhao2017topic}.

% Table generated by Excel2LaTeX from sheet 'Sheet1'
\begin{table*}[htbp]
	\scriptsize
	\centering
	\begin{tabular}{|c|ccc|ccc|ccc|ccc|}
		\hline
		\textbf{Model} & \multicolumn{3}{c|}{\textbf{MR}} & \multicolumn{3}{c|}{\textbf{SUBJ}} & \multicolumn{3}{c|}{\textbf{CR}} & \multicolumn{3}{c|}{\textbf{TREC}} \\
		\hline
		CNN   & \multicolumn{3}{c|}{81.5} & \multicolumn{3}{c|}{93.4} & \multicolumn{3}{c|}{85.0} & \multicolumn{3}{c|}{93.6} \\
		Dep-CNN & \multicolumn{3}{c|}{81.9} & \multicolumn{3}{c|}{-} & \multicolumn{3}{c|}{-} & \multicolumn{3}{c|}{95.4} \\
		DSCNN & \multicolumn{3}{c|}{82.2} & \multicolumn{3}{c|}{93.2} & \multicolumn{3}{c|}{-} & \multicolumn{3}{c|}{\textbf{95.6}} \\
		\hline
		% RNN   & \multicolumn{3}{c|}{77.2} & \multicolumn{3}{c|}{90.9} & \multicolumn{3}{c|}{71.8} & \multicolumn{3}{c|}{83.8} \\
		% LSTM  & \multicolumn{3}{c|}{79.5} & \multicolumn{3}{c|}{93.3} & \multicolumn{3}{c|}{80.4} & \multicolumn{3}{c|}{89.4} \\
		% GRU   & \multicolumn{3}{c|}{80.5} & \multicolumn{3}{c|}{93.5} & \multicolumn{3}{c|}{82.1} & \multicolumn{3}{c|}{91.8} \\
		% \hline
		AdaSent & \multicolumn{3}{c|}{\textbf{83.1}} & \multicolumn{3}{c|}{\underline{\textbf{95.5}}} & \multicolumn{3}{c|}{86.3} & \multicolumn{3}{c|}{92.4} \\
		\hline
		\hline
		\textbf{C = Topic} & \multicolumn{1}{c}{word} & \multicolumn{1}{c}{sent} & \multicolumn{1}{c|}{ens} & \multicolumn{1}{c}{word} & \multicolumn{1}{c}{sent} & \multicolumn{1}{c|}{ens} & \multicolumn{1}{c}{word} & \multicolumn{1}{c}{sent} & \multicolumn{1}{c|}{ens} & \multicolumn{1}{c}{word} & \multicolumn{1}{c}{sent} & \multicolumn{1}{c|}{ens} \\
		\hline
		TopCNN & \makecell{81.7\\(+0.2)} & \textcolor[rgb]{ 1,  0,  0}{\makecell{81.3\\(-0.2)}} & \makecell{83.0\\(+1.5)} & \textcolor[rgb]{ 1,  0,  0}{\makecell{93.4\\(+0.0)}} & \textcolor[rgb]{ 1,  0,  0}{\makecell{93.4\\(+0.0)}} & \makecell{95.0\\(+1.6)} & \textcolor[rgb]{ 1,  0,  0}{\makecell{84.9\\(-0.1)}} & \textcolor[rgb]{ 1,  0,  0}{\makecell{84.8\\(-0.2)}} & \textbf{\makecell{86.4\\(+1.4)}} & \textcolor[rgb]{ 1,  0,  0}{\makecell{92.5\\(-1.1)}} & \textcolor[rgb]{ 1,  0,  0}{\makecell{92.0\\(-1.6)}} & \makecell{94.0\\(+0.4)} \\
		\hline
		\hline
		\textbf{C = Trans} & \multicolumn{1}{c}{N=1} & \multicolumn{1}{c}{N=10} & \multicolumn{1}{c|}{ens} & \multicolumn{1}{c}{N=1} & \multicolumn{1}{c}{N=10} & \multicolumn{1}{c|}{ens} & \multicolumn{1}{c}{N=1} & \multicolumn{1}{c}{N=10} & \multicolumn{1}{c|}{ens} & \multicolumn{1}{c}{N=1} & \multicolumn{1}{c}{N=10} & \multicolumn{1}{c|}{ens} \\
		\hline
		CNN+B1 & \makecell{81.9\\(+0.4)} & \textcolor[rgb]{ 1,  0,  0}{\makecell{81.4\\(-0.1)}} & \makecell{82.6\\(+1.1)} & \makecell{94.6\\(+1.2)} & \makecell{93.8\\(+0.4)} & \makecell{94.9\\(+1.5)} & \makecell{86.2\\(+1.2)} & \makecell{85.9\\(+0.9)} & \makecell{86.7\\(+1.7)} & \makecell{95.4\\(+1.8)} & \makecell{95.0\\(+1.4)} & \makecell{96.4\\(+3.0)} \\
		CNN+B2 & \makecell{82.1\\(+0.6)} & \makecell{82.1\\(+0.6)} & \makecell{82.2\\(+0.7)} & \makecell{94.6\\(+1.2)} & \makecell{94.0\\(+0.6)} & \makecell{94.8\\(+1.4)} & \makecell{86.1\\(+1.1)} & \makecell{86.3\\(+1.3)} & \makecell{86.6\\(+1.6)} & \makecell{95.4\\(+1.8)} & \makecell{95.2\\(+1.6)} & \makecell{96.4\\(+3.0)} \\
		CNN+MCFA & \makecell{82.3\\(+0.8)} & \makecell{82.7\\(+1.2)} & \textbf{\makecell{\underline{83.2}\\\underline{(+1.7)}}} & \makecell{94.7\\(+1.3)} & \makecell{94.8\\(+1.4)} & \textbf{\makecell{95.2\\(+1.8)}} & \makecell{87.6\\(+2.6)} & \makecell{88.6\\(+3.6)} & \textbf{\makecell{\underline{89.4}\\\underline{(+4.4)}}} & \makecell{95.4\\(+1.8)} & \makecell{96.0\\(+2.4)} & \textbf{\makecell{\underline{96.8}\\\underline{(+3.4)}}} \\
		\hline
	\end{tabular}%
	\caption{Classification accuracies of competing models. \textbf{C} refers to the additional context, $N$ refers to the number of translations. In TopCNN, word refers to using word-specific topic while sentence refers to using sentence-specific topic. 
		%The first seven models are models without additional context. The rest of the models are with additional context.
		Accuracies colored \textcolor[rgb]{1,0,0}{red} are accuracies that perform worse than CNN. Previous state of the art results and the results of our best model are \textbf{bold-faced}. The winning result is \underline{underlined}. The number inside the parenthesis indicates the increase from the base model, CNN.}
	\label{tab:result}%
\end{table*}%

\subsection{Results and Discussion}

We report the classification accuracy of the competing models in Table \ref{tab:result}. We show that CNN+MCFA achieves state of the art performance on three of the four data sets and performs competitively on one data set. When $N=1$, MCFA increases the performance of a normal CNN from $85.0$ to $87.6$, beating the current state of the art on the CR data set. When $N=10$, MCFA additionally beats the state of the art on the TREC data set. Finally, our ensemble classifier additionally outperforms all competing models on the MR data set. We emphasize that we only use the basic CNN as our sentence encoder for our experiments, yet still achieve state of the art performance on most data sets. Hence, MCFA is successful in effectively using translations as additional context to improve the performance of the classifier.

We compare our model (CNN+MCFA) and the baselines discussed above (CNN+B1, CNN+B2). On all settings, our model outperforms the baselines. When $N=10$, the performance of our model increases over the performance when $N=1$, however the performance of CNN+B1 decreases when compared to the performance when $N=1$. We also show the accuracies of the worst classifiers when $N=1$ in Table \ref{tab:minresult}. On all data sets except SUBJ, the accuracy of CNN+B1 decreases from the base CNN accuracy, while the accuracy of our model always improves from the base CNN accuracy.
%We explain that these phenomena are due to the increase of noisy vectors due to wrong translations.
This is resolved by CNN+B2 by applying L2 regularization, however the increase in performance is marginal. 

\begin{table}[t]
	\scriptsize
  \centering
    \begin{tabular}{|c|c|c|c|c|}
    \hline
    \textbf{Model} & {\textbf{MR}} & \textbf{SUBJ} & \textbf{CR} & {\textbf{TREC}} \\
    \hline
    CNN   & 81.5  & 93.4  & 85.0    & 93.6 \\
    \hline
    CNN+B1 & \textcolor[rgb]{ 1,  0,  0}{81.4} & 94.2  & \textcolor[rgb]{ 1,  0,  0}{83.8} & \textcolor[rgb]{ 1,  0,  0}{93.0} \\
    CNN+B2 & 81.7  & 94.2  & \textcolor[rgb]{ 1,  0,  0}{84.0} & \textcolor[rgb]{ 1,  0,  0}{93.2} \\
    CNN+MCFA & 81.8  & 94.4  & 85.8  & 94.2 \\
    \hline
    \end{tabular}%
  \caption{Accuracies of the worst CNN+translation classifiers when $N=1$. Accuracies less than CNN accuracies are highlighted in \textcolor[rgb]{1,0,0}{red}.}
  \label{tab:minresult}%
\end{table}%

We also compare two different kinds of additional context: topics (TopCNN) and translations (CNN+B1, CNN+B2, CNN+MCFA). Overall, we conclude that translations are better additional contexts than topics. When using a single context (i.e. TopCNN$_\text{word}$, TopCNN$_\text{sent}$, and our models when $N=1$), translations always outperform topics even when using the baseline methods. Using topics as additional context also decreases the performance of the CNN classifier on most data sets, giving an adverse effect to the CNN classifier.

%Further investigation on the sentence vectors explains how translations as additional context is effective. We visualize sentence vectors on the TREC data set using PCA and show visualizations of vectors in three languages in Figure \ref{fig:trecvecs}. We see that each language is better at classifying different classes. Specifically, English, Arabic, and French sentence vectors are respectively better at classifying the orange, green, and red classes, which are highlighted in the Figure. This means that translation as context gives a different perspective on viewing the sentence and thus may provide better features in classification.

\section{Model Interpretation}

\begin{figure}[t]
	\centering
	\begin{subfigure}{0.4\textwidth}
		\centering
		\includegraphics[width=0.85\textwidth]{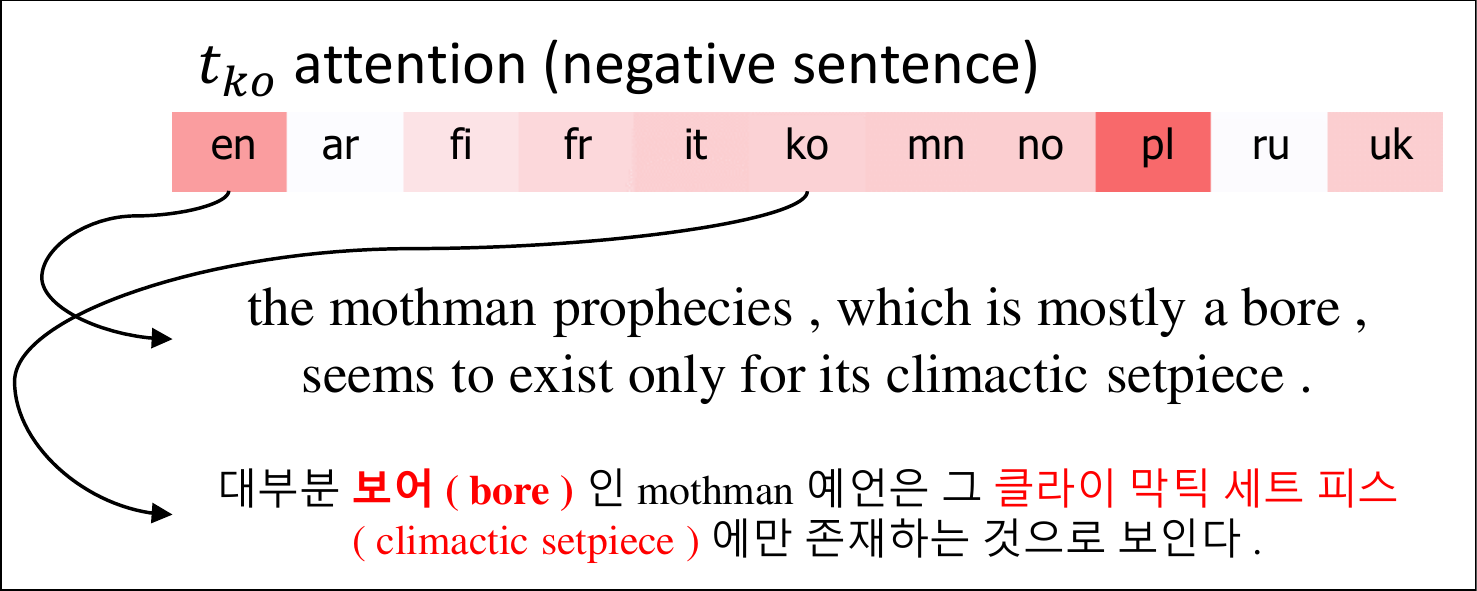}
		\caption{Example where English attention weight is larger}
	\end{subfigure}
	\begin{subfigure}{0.4\textwidth}
		\centering
		\includegraphics[width=0.85\textwidth]{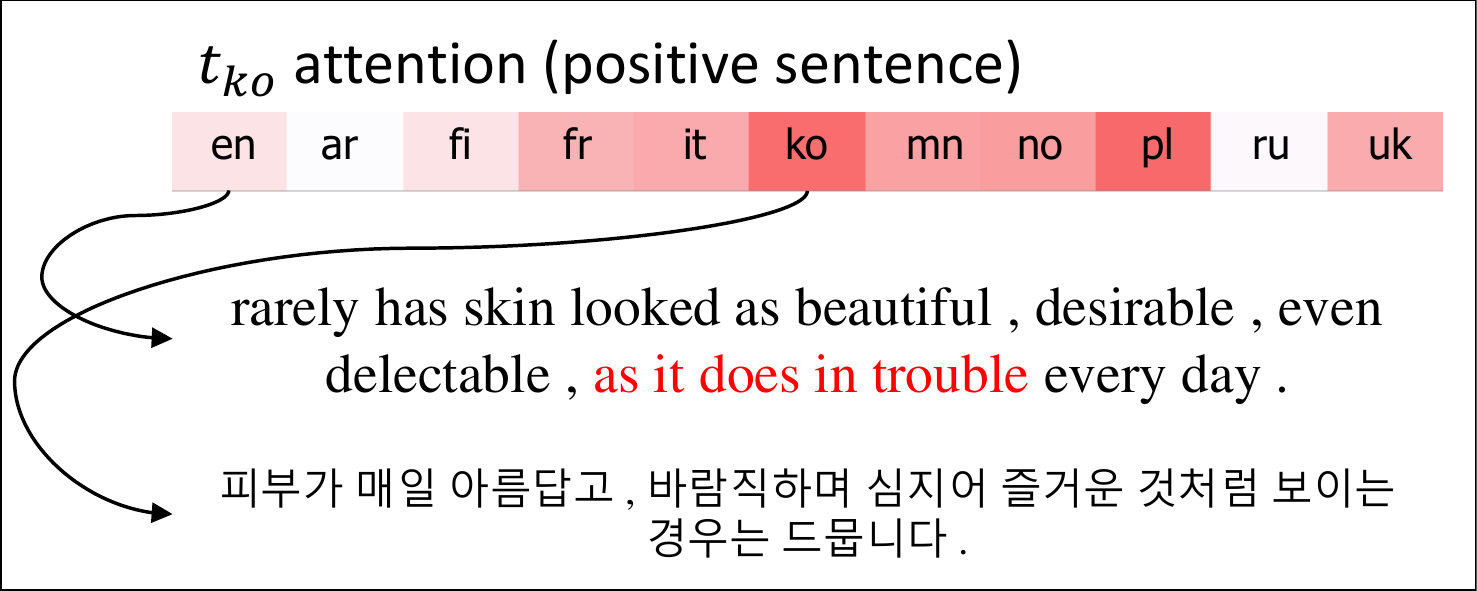}
		\caption{Example where Korean attention weight is larger}
	\end{subfigure}
	\caption{Attention weights of example Korean sentences from the MR data set. The red color fill represents the attention weights given to each sentence. The darker the fill, the larger the attention weight.}
	\label{fig:attention}
\end{figure}

\begin{CJK}{UTF8}{}
 \CJKfamily{mj}
\begin{table}[!ht]
	\scriptsize
    \centering
    \begin{subtable}{0.47\textwidth}
        \centering
        \begin{tabularx}{\textwidth}{X}
            Original sentence: \\
            \textit{skip this turd and pick your nose instead because you're sure to get more out of the latter experience .} \\ \hline
            Korean translation: \\
            \textit{\textcolor[rgb]{1,0,0}{후자의 경험에서 더 많은 것을 얻으려면} 이 웅덩이를 건너 뛰고 \textcolor[rgb]{1,0,0}{코를 골라야합니다} .} \\ \hline
            Human re-translation: \\ 
            \textit{In order to get more from the latter experience , you need to skip this puddle and choose your nose .} \\ \hline
            \textbf{Self Usability: 0.3958} \\
        \end{tabularx}
        \caption{Low self usability example}
    \end{subtable}
    \begin{subtable}{0.47\textwidth}
        \centering
        \begin{tabularx}{\textwidth}{X}
            Original sentence: \\
            \textit{michael moore's latest documentary about america's thirst for violence is his best film yet . . .} \\ \hline
            Korean translation: \\
            \textit{마이클 무어 ( Michael Moore ) 의 최근 미국 다큐멘터리 \textcolor[rgb]{1,0,0}{`` 폭력 장면 ''} 은 그의 최고의 영화 다 . . .} \\ \hline
            Human re-translation: \\
            \textit{Michael Moore's latest American documentary `` Violent Scene '' is his best film yet . . .} \\ \hline
            \textbf{Self Usability: 1.0000} \\
        \end{tabularx}
        \caption{High self usability example}
    \end{subtable}
    \caption{Two examples of self usability of Korean sentences from the MR data set. Texts colored in \textcolor[rgb]{1,0,0}{red} are mistranslated texts.}
    \label{tab:usability}
\end{table}
\end{CJK}

%In the next sections, we look at the different parts of MCFA
%: the self usability module, the relative usability module, and the vector fixing module,
%and provide visualizations
%examples and explanations
%for interpretation.

%\paragraph{Self usability}

We first provide examples shown in Table \ref{tab:usability} on how the self usability module determines the score of sentences. In the first example, it is hard to classify whether the translated sentence is positive or negative, thus it is given a low self usability score. In the second example, although the sentence contains mistranslations, these are minimal and may actually help the classifier by telling it that \textit{thirst for violence} is not a negative phrase. Thus, it is given a high self usability score.

%\paragraph{Attention weights}

\begin{figure}[!t]
    \centering
    \includegraphics[width=0.45\textwidth]{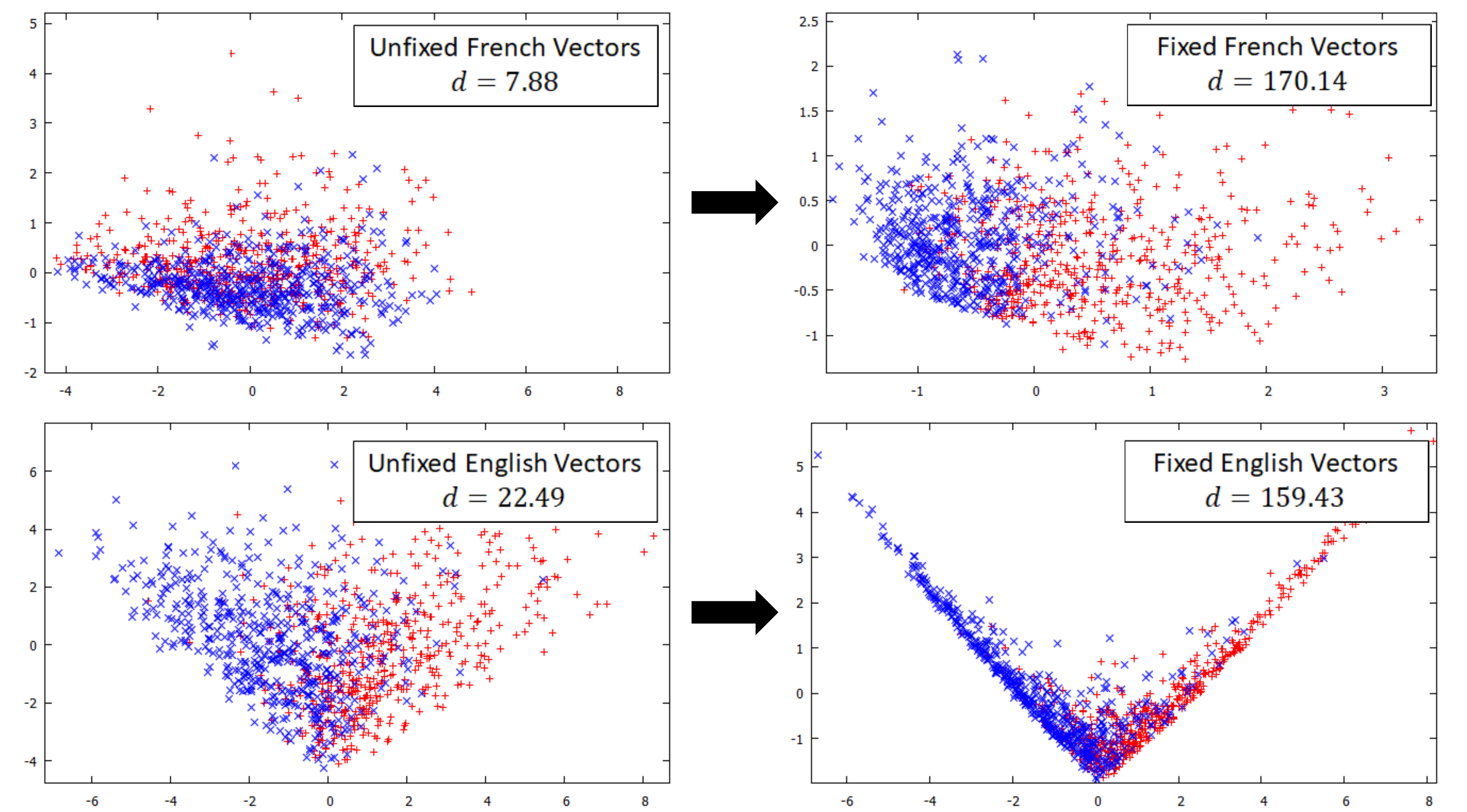}
    % \begin{subfigure}{0.45\textwidth}
    %     \centering
    %     \includegraphics[width=\textwidth]{ar2}
    %     \caption{Arabic sentence vectors}
    % \end{subfigure}
    % \begin{subfigure}{0.45\textwidth}
    %     \centering
    %     \includegraphics[width=\textwidth]{fr2}
    %     \caption{French sentence vectors}
    % \end{subfigure}
    % \begin{subfigure}{0.45\textwidth}
    %     \centering
    %     \includegraphics[width=\textwidth]{ko2}
    %     \caption{Korean sentence vectors}
    % \end{subfigure}
    % \begin{subfigure}{0.45\textwidth}
    %     \centering
    %     \includegraphics[width=\textwidth]{en2}
    %     \caption{English sentence vectors}
    % \end{subfigure}
    \caption{PCA visualization of unaltered (left) and altered (right) vectors of %one CV
    %cross validation 
    %test set of
    the MR data set. $d$ is the Mahalanobis distance between two class clusters.}
    \label{fig:mrvecsbefaf}
\end{figure}

\begin{table}[t]
	\scriptsize
  \centering
  \def\arraystretch{0.7}
    \begin{tabular}{|m{0.06\textwidth}|m{0.36\textwidth}|} %|c|}
    \hline
    \textbf{Sentence} & may take its sweet time to get wherever it's going, but if you have the patience for it, you won't feel like it's wasted yours. \\% & \textbf{Similarity}\\
    \hline
    \textbf{NN \newline (Unaltered)} & you know that ten bucks you'd spend on a ticket? just send it to cranky. we don't get paid enough to sit through crap like this. \\%& 0.682 \\
    \hline
    \textbf{NN \newline (altered)} & what might have been readily dismissed as the tiresome rant of an aging filmmaker still thumbing his nose at convention takes a surprising, subtle turn at the midway point. \\% & 0.768\\
    \hline
    \end{tabular}%
    \vspace*{0.1cm}
    \begin{tabular}{|m{0.06\textwidth}|m{0.36\textwidth}|} %|c|}
    \hline
    \textbf{Sentence} & every nanosecond of the new guy reminds 
    %you 
    that you could be doing something else 
    %far 
    more pleasurable. 
    %something 
    like scrubbing the toilet. 
    %or 
    emptying rat traps. or doing last year's taxes with your ex-wife. \\%& \textbf{Similarity}\\
    \hline
    \textbf{NN \newline (Unaltered)} & in the new release of cinema paradiso, the tale has turned from sweet to bittersweet, and when the tears come during that final, beautiful scene, they finally feel absolutely earned. \\%& 0.734 \\
    \hline
    \textbf{NN \newline (altered)} & after 
    %several
     scenes of 
     % this
    % tacky 
    nonsense, you'll be wistful for the testosterone-charged wizardry of jerry bruckheimer productions, especially because half past dead is like the rock on 
    %a
     walmart budget. \\%& 0.735\\
    \hline
    \end{tabular}%
  \caption{Two example sentences, from English (first) and Korean (second) vector spaces, and their nearest neighbors (NN) on both the unaltered and altered vector spaces. We only show the original English sentences for the Korean example for conciseness.}
  \label{tab:vecfixexample}%
\end{table}%

Figure \ref{fig:attention} shows two data instance examples where we show the attention weights given to the other contexts when fixing a Korean sentence. The larger the attention weight is, the more the context is used to fix the Korean sentence. In the first example, the Korean sentence contains translation errors; especially, the words \textit{bore} and \textit{climactic setpiece} were not translated and were only spelled using the Korean alphabet. In this example, the English attention weight is larger than the Korean attention weight. In the second example, the Korean sentence correctly translates all parts of the English sentence, except for the phrase \textit{as it does in trouble}. However, this phrase is not necessary to classify the sentence correctly, and may induce possible vagueness %classification 
because of the word \textit{trouble}. Thus, the Korean attention weight is larger.

%\paragraph{Vector fixers}

%We show the effectiveness of the vector fixing module in fixing the context vectors. 
Figure \ref{fig:mrvecsbefaf} shows the PCA visualization of the unaltered and the altered vectors of four different languages. In the first 
%and second 
example, the unaltered 
%Arabic and French 
sentence vectors are mostly in the middle of the vector space, making it hard to draw a boundary between
%positive and negative
the two
examples. After the fixing, the boundary is much clearer.
\begin{comment}
In the second example, it is impossible to distinguish the classes as
%because %most of 
the vectors 
%in the unaltered space 
are overlapping with each other. After the fix,
%fixing the vectors,
the blue-class and red-class vectors moved to the opposite sides,
%upwards and the red-class vectors moved downwards, 
making it easier to tell and differentiate the two classes. 
\end{comment}
We also show the English sentence vectors in the second example. Even without fixing the unaltered English sentence vectors, it is easy to distinguish both classes. %However, a
After the fix, the sentence vectors in the middle of the space are moved,
%to their corresponding sides, 
making the distinction more obvious and clearer. We also provide quantitative evidence by showing that the Mahalanobis distance between the two classes in the altered vectors are significantly farther than that of the unaltered vectors.

We also show two examples sentences from English and Korean vector spaces and their corresponding nearest neighbors on both the unaltered and altered vector spaces in Table \ref{tab:vecfixexample}. In the first example, the unaltered vector focuses on the meaning of \textit{``wasted yours''} in the sentence, which puts it near sentences regarding wasted time or money. After fixing, the sentence vector focuses its meaning on the slow yet worth-the-wait pace of the movie, thus moving it closer to the correct vectors. In the second example, all three sentences have highly descriptive tones, however, the nearest neighbor on the altered space is hyperbolically negative, comparing the movie to a description unrelated to the movie itself.

\section{Related Work}

%\paragraph{Sentence classification with additional context}

One way to improve the performance of a sentence classifier is to introduce new context. Common and obvious kinds of context are the neighboring sentences of the sentence \cite{lin2015hierarchical}, and the document where the sentence belongs \cite{huang2012improving}. 
%Another possible type of context is the topic \cite{amiri2016learning} of the sentence. 
Topics of the words in the sentence induced by a topic model
%Latent Dirichlet Allocation (LDA) topic model \cite{blei2012probabilistic}, combined with sentence topics, 
were also used as contexts \cite{zhao2017topic}.
In this paper, we introduce yet another type of additional context, sentence translations, which to the best of our knowledge have not been used previously.

%\paragraph{Transfer learning using machine translation}

Sentence encoders trained from neural machine translation (NMT) systems were also used for transfer learning \cite{hill2016learning}. 
%The effects of transfer learning using NMT encoders from a variety of source domains were first studied to semantic similarity tasks \cite{hill2016learning}. 
\cite{hill2017representational} 
%further 
demonstrated that altered-length sentence vectors from NMT encoders outperform sentence vectors from monolingual encoders on semantic similarity tasks. Recent work used representation of each word in the sentence to create a sentence representation suitable for multiple NLP tasks \cite{mccann2017learned}.
Our work shares the commonality of using NMT for another task, but instead of using NMT to encode our sentences, we 
%but rather 
use it to translate the sentences into new contexts.
%These contexts are then encoded into sentences using sentence encoders such as CNN \cite{Kim2014ConvolutionalNN}.

%\paragraph{Dataset augmentation for sentence classification}

Increasing the number of data instances of the training set has also been explored to improve the performance of a classifier. 
%Past methods include incremental learning where the model is increasingly trained using automatically labelled data \cite{yan2009svm}.
Recent methods include the usage of thesaurus \cite{zhang2015character}, paraphrases \cite{fu2014improving}, among others.
%and context re-ordering \cite{pan2016expanding}. 
These simple variation techniques are preferred because they are found to be very effective despite their simplicity.
%This line of work differs from ours in such a way that dataset augmentation increases the dataset \textit{vertically} while additional contexts increases the dataset \textit{horizontally}.
Our work similarly augments training data, not by adding data instances (vertical augmentation), but rather by adding more context (horizontal augmentation). Though the paraphrase of $p$ can be alternatively used as an augmented context, this could not leverage the added semantics coming from another language, as discussed in Section~\ref{sec:intro}. 

%\paragraph{Cross-language text classification} Another similar area is on cross-language text classification, where a classifier created using a data set in one language is used to effectively classify texts in another language \cite{}. 

\section{Conclusion}

This paper investigates the use of translations as better additional contexts for sentence classification. To answer the problem on mistranslations, we propose 
%a neural attention-based solution called
multiple context fixing attachment (MCFA) to fix the context vectors using other context vectors. We show that our method improves the classification performance and achieves state-of-the-art performance on multiple data sets. %We also provide qualitative analysis on how our method is effective on fixing noisy context vectors and how languages help differently depending on the dataset. 
%We provide comprehensive details of the classification results and visualizations on a separate supplementary material \footnote{See attached supplementary material.}.
%MCFA uses an advantageous approach to fix the vectors mainly because it is extensible to any kind of sentence encoding methods since it is non-invasive, i.e. it is attached after the encoding of the sentence. It is interesting to explore non-invasive attachment techniques on other types of sentence encoder for other types of NLP tasks.
In our future work, we plan to use and extend our model to other complex NLP tasks.
%such as 
%natural 
%language inference and question answering.

\section*{Acknowledgements}

This work was supported by Microsoft Research Asia and the ICT R\&D program of MSIT/IITP.
[2017-0-01778, Development of Explainable Human-level Deep Machine Learning Inference Framework]

\bibliographystyle{named}
\bibliography{ijcai18}

\begin{thebibliography}{}

\bibitem[\protect\citeauthoryear{Bahdanau \bgroup \em et al.\egroup
  }{2014}]{bahdanau2014neural}
Dzmitry Bahdanau, Kyunghyun Cho, and Yoshua Bengio.
\newblock Neural machine translation by jointly learning to align and
  translate.
\newblock {\em arXiv preprint arXiv:1409.0473}, 2014.

\bibitem[\protect\citeauthoryear{Bojanowski \bgroup \em et al.\egroup
  }{2016}]{bojanowski2016enriching}
Piotr Bojanowski, Edouard Grave, Armand Joulin, and Tomas Mikolov.
\newblock Enriching word vectors with subword information.
\newblock {\em arXiv preprint arXiv:1607.04606}, 2016.

\bibitem[\protect\citeauthoryear{Collobert \bgroup \em et al.\egroup
  }{2011}]{collobert2011natural}
Ronan Collobert, Jason Weston, L{\'e}on Bottou, Michael Karlen, Koray
  Kavukcuoglu, and Pavel Kuksa.
\newblock Natural language processing (almost) from scratch.
\newblock {\em Journal of Machine Learning Research}, 12(Aug):2493--2537, 2011.

\bibitem[\protect\citeauthoryear{Fu \bgroup \em et al.\egroup
  }{2014}]{fu2014improving}
Guohong Fu, Yu~He, Jiaying Song, and Chaoyue Wang.
\newblock Improving chinese sentence polarity classification via opinion
  paraphrasing.
\newblock {\em CLP 2014}, page~35, 2014.

\bibitem[\protect\citeauthoryear{He \bgroup \em et al.\egroup
  }{2016}]{he2016deep}
Kaiming He, Xiangyu Zhang, Shaoqing Ren, and Jian Sun.
\newblock Deep residual learning for image recognition.
\newblock In {\em CVPR}, pages 770--778, 2016.

\bibitem[\protect\citeauthoryear{Hill \bgroup \em et al.\egroup
  }{2016}]{hill2016learning}
Felix Hill, Kyunghyun Cho, and Anna Korhonen.
\newblock Learning distributed representations of sentences from unlabelled
  data.
\newblock {\em arXiv preprint arXiv:1602.03483}, 2016.

\bibitem[\protect\citeauthoryear{Hill \bgroup \em et al.\egroup
  }{2017}]{hill2017representational}
Felix Hill, Kyunghyun Cho, S{\'e}bastien Jean, and Yoshua Bengio.
\newblock The representational geometry of word meanings acquired by neural
  machine translation models.
\newblock {\em Machine Translation}, pages 1--16, 2017.

\bibitem[\protect\citeauthoryear{Hu and Liu}{2004}]{hu2004mining}
Minqing Hu and Bing Liu.
\newblock Mining and summarizing customer reviews.
\newblock In {\em SIGKDD}, pages 168--177. ACM, 2004.

\bibitem[\protect\citeauthoryear{Huang \bgroup \em et al.\egroup
  }{2012}]{huang2012improving}
Eric~H Huang, Richard Socher, Christopher~D Manning, and Andrew~Y Ng.
\newblock Improving word representations via global context and multiple word
  prototypes.
\newblock In {\em ACL}, pages 873--882. Association for Computational
  Linguistics, 2012.

\bibitem[\protect\citeauthoryear{Johnson \bgroup \em et al.\egroup
  }{2016}]{johnson2016google}
Melvin Johnson, Mike Schuster, Quoc~V Le, Maxim Krikun, Yonghui Wu, Zhifeng
  Chen, Nikhil Thorat, Fernanda Vi{\'e}gas, Martin Wattenberg, Greg Corrado,
  et~al.
\newblock Google's multilingual neural machine translation system: enabling
  zero-shot translation.
\newblock {\em arXiv preprint arXiv:1611.04558}, 2016.

\bibitem[\protect\citeauthoryear{Joulin \bgroup \em et al.\egroup
  }{2017}]{Joulin2017BagOT}
Armand Joulin, Edouard Grave, Piotr Bojanowski, and Tomas Mikolov.
\newblock Bag of tricks for efficient text classification.
\newblock In {\em EACL}, 2017.

\bibitem[\protect\citeauthoryear{Kim}{2014}]{Kim2014ConvolutionalNN}
Yoon Kim.
\newblock Convolutional neural networks for sentence classification.
\newblock In {\em EMNLP}, 2014.

\bibitem[\protect\citeauthoryear{Li and Roth}{2002}]{li2002learning}
Xin Li and Dan Roth.
\newblock Learning question classifiers.
\newblock In {\em COLING}, pages 1--7. Association for Computational
  Linguistics, 2002.

\bibitem[\protect\citeauthoryear{Lin \bgroup \em et al.\egroup
  }{2015}]{lin2015hierarchical}
Rui Lin, Shujie Liu, Muyun Yang, Mu~Li, Ming Zhou, and Sheng Li.
\newblock Hierarchical recurrent neural network for document modeling.
\newblock In {\em EMNLP}, pages 899--907, 2015.

\bibitem[\protect\citeauthoryear{Ma \bgroup \em et al.\egroup
  }{2015}]{ma2015dependency}
Mingbo Ma, Liang Huang, Bing Xiang, and Bowen Zhou.
\newblock Dependency-based convolutional neural networks for sentence
  embedding.
\newblock {\em arXiv preprint arXiv:1507.01839}, 2015.

\bibitem[\protect\citeauthoryear{McCann \bgroup \em et al.\egroup
  }{2017}]{mccann2017learned}
Bryan McCann, James Bradbury, Caiming Xiong, and Richard Socher.
\newblock Learned in translation: Contextualized word vectors.
\newblock {\em arXiv preprint arXiv:1708.00107}, 2017.

\bibitem[\protect\citeauthoryear{Mimno \bgroup \em et al.\egroup
  }{2011}]{mimno2011optimizing}
David Mimno, Hanna~M Wallach, Edmund Talley, Miriam Leenders, and Andrew
  McCallum.
\newblock Optimizing semantic coherence in topic models.
\newblock In {\em EMNLP}, pages 262--272. Association for Computational
  Linguistics, 2011.

\bibitem[\protect\citeauthoryear{Pang and Lee}{2004}]{pang2004sentimental}
Bo~Pang and Lillian Lee.
\newblock A sentimental education: Sentiment analysis using subjectivity
  summarization based on minimum cuts.
\newblock In {\em ACL}, page 271. Association for Computational Linguistics,
  2004.

\bibitem[\protect\citeauthoryear{Pang and Lee}{2005}]{pang2005seeing}
Bo~Pang and Lillian Lee.
\newblock Seeing stars: Exploiting class relationships for sentiment
  categorization with respect to rating scales.
\newblock In {\em ACL}, pages 115--124. Association for Computational
  Linguistics, 2005.

\bibitem[\protect\citeauthoryear{Pang and Lee}{2007}]{Pang2007OpinionMA}
Bo~Pang and Lillian Lee.
\newblock Opinion mining and sentiment analysis.
\newblock {\em Foundations and Trends in Information Retrieval}, 2:1--135,
  2007.

\bibitem[\protect\citeauthoryear{Srivastava \bgroup \em et al.\egroup
  }{2014}]{srivastava2014dropout}
Nitish Srivastava, Geoffrey~E Hinton, Alex Krizhevsky, Ilya Sutskever, and
  Ruslan Salakhutdinov.
\newblock Dropout: a simple way to prevent neural networks from overfitting.
\newblock {\em Journal of machine learning research}, 15(1):1929--1958, 2014.

\bibitem[\protect\citeauthoryear{Zhang \bgroup \em et al.\egroup
  }{2015}]{zhang2015character}
Xiang Zhang, Junbo Zhao, and Yann LeCun.
\newblock Character-level convolutional networks for text classification.
\newblock In {\em NIPS}, pages 649--657, 2015.

\bibitem[\protect\citeauthoryear{Zhang \bgroup \em et al.\egroup
  }{2016}]{zhang2016dependency}
Rui Zhang, Honglak Lee, and Dragomir Radev.
\newblock Dependency sensitive convolutional neural networks for modeling
  sentences and documents.
\newblock {\em arXiv preprint arXiv:1611.02361}, 2016.

\bibitem[\protect\citeauthoryear{Zhao \bgroup \em et al.\egroup
  }{2015}]{zhao2015self}
Han Zhao, Zhengdong Lu, and Pascal Poupart.
\newblock Self-adaptive hierarchical sentence model.
\newblock In {\em IJCAI}, pages 4069--4076, 2015.

\bibitem[\protect\citeauthoryear{Zhao \bgroup \em et al.\egroup
  }{2017}]{zhao2017topic}
Rui Zhao, Kezhi Mao, Rui Zhao, and Kezhi Mao.
\newblock Topic-aware deep compositional models for sentence classification.
\newblock {\em IEEE/ACM Transactions on Audio, Speech and Language Processing},
  25(2):248--260, 2017.

\end{thebibliography}

\end{document}